\documentclass[11pt,a4paper]{article}
\usepackage[hyphens]{url}
\usepackage[hyperref]{emnlp2020}
\usepackage{times}
\usepackage{latexsym}
\usepackage{color}
\usepackage{breakurl}
\usepackage{graphicx}
\usepackage{booktabs}
\usepackage{subcaption}
\usepackage{geometry}
\usepackage{multirow}

\usepackage{microtype}

\aclfinalcopy 
\def\aclpaperid{1752} 

\title{Participatory Research for Low-resourced Machine Translation: \\ A Case Study in African Languages}

\author{
\begin{minipage}[t]{\textwidth}
\centering
\normalsize
$\forall$\thanks{$\forall$ to represent the whole Masakhane community.}\,,
Wilhelmina Nekoto$^{1}$,
Vukosi Marivate$^{2}$,
Tshinondiwa Matsila$^{1}$,
Timi Fasubaa$^{3}$,\\
Tajudeen Kolawole$^{4}$,
Taiwo Fagbohungbe$^{5}$,
Solomon Oluwole Akinola$^{6}$,\\
Shamsuddeen Hassan Muhammad$^{7,39}$,
Salomon Kabongo$^{4}$,
Salomey Osei$^{4}$,\\
Sackey Freshia$^{8}$,
Rubungo Andre Niyongabo$^{9}$,
Ricky Macharm$^{10}$,
Perez Ogayo$^{11}$,
Orevaoghene Ahia$^{12}$,
Musie Meressa$^{13}$,
Mofe Adeyemi$^{14}$,
Masabata Mokgesi-Selinga$^{15}$,
Lawrence Okegbemi$^{5}$,
Laura Jane Martinus$^{16}$,
Kolawole Tajudeen$^{4}$,
Kevin Degila$^{17}$,\\
Kelechi Ogueji$^{12}$,
Kathleen Siminyu$^{18}$,
Julia Kreutzer$^{19}$,
Jason Webster$^{20}$,\\
Jamiil Toure Ali$^{1}$,
Jade Abbott$^{21}$,
Iroro Orife$^{3}$,
Ignatius Ezeani$^{38}$,\\
Idris Abdulkabir Dangana$^{23,7}$,
Herman Kamper$^{24}$,
Hady Elsahar$^{25}$,
Goodness Duru$^{26}$,\\
Ghollah Kioko$^{27}$,
Espoir Murhabazi$^{1}$,
Elan van Biljon$^{12,24}$,
Daniel Whitenack$^{28}$,\\
Christopher Onyefuluchi$^{29}$,
Chris Emezue$^{40}$,
Bonaventure Dossou$^{31}$,
Blessing Sibanda$^{32}$,
Blessing Itoro Bassey$^{4}$,
Ayodele Olabiyi$^{33}$,
Arshath Ramkilowan$^{34}$,
Alp \"Oktem$^{35}$,
Adewale Akinfaderin$^{36}$,
Abdallah Bashir$^{37}$ \\

{\footnotesize \normalfont 
$^*$Masakhane, Africa
$^{1}$Independent,
$^{2}$University of Pretoria,
$^{3}$Niger-Volta LTI,\\
$^{4}$African Masters in Machine Intelligence,
$^{5}$Federal University of Technology, Akure,
$^{6}$University of Johannesburg,\\
$^{7}$Bayero University, Kano,
$^{8}$Jomo Kenyatta University of Agriculture and Technology,
$^{9}$UESTC,\\
$^{10}$Siseng Consulting Ltd,
$^{11}$African Leadership University,
$^{12}$InstaDeep Ltd,\\
$^{13}$Sapienza University of Rome,
$^{14}$Udacity,
$^{15}$Parliament, Republic of South Africa,\\
$^{16}$Explore Data Science Academy,
$^{17}$UCD, Konta,
$^{18}$AI4Dev,
$^{19}$Google Research,
$^{20}$Percept,
$^{21}$Retro Rabbit,\\
$^{23}$Di-Hub,
$^{24}$Stellenbosch University,
$^{25}$Naver Labs Europe,
$^{26}$Retina, AI,
$^{27}$Lori Systems,
$^{28}$SIL International,\\
$^{29}$Federal College of Dental Technology and Therapy, Enugu,
$^{31}$Jacobs University,\\
$^{32}$Namibia University of Science and Technology,
$^{33}$Data Science Nigeria,
$^{34}$Praekelt Consulting,\\
$^{35}$Translators without Borders,
$^{36}$Amazon,
$^{37}$Max Planck Institute for Informartics, University of Saarland,\\
$^{38}$Lancaster University, 
$^{39}$University of Porto, 
$^{40}$Technical University of Munich\\
} 

{\small \normalfont \texttt{masakhane-nlp@googlegroups.com}}
\end{minipage}
}

\date{}

\begin{document}

\maketitle
\begin{abstract}
Research in NLP lacks geographic diversity, and the question of how NLP can be scaled to low-resourced languages has not yet been adequately solved. ``Low-resourced''-ness is a complex problem going beyond data availability and reflects systemic problems in society.

In this paper, we focus on the task of Machine Translation (MT), that plays a crucial role for information accessibility and communication worldwide. Despite immense improvements in MT over the past decade, MT is centered around a few high-resourced languages. 

As MT researchers cannot solve the problem of low-resourcedness alone, we propose participatory research as a means to involve all necessary agents required in the MT development process. We demonstrate the feasibility and scalability of participatory research with a case study on MT for African languages. Its implementation leads to a collection of novel translation datasets, MT benchmarks for over 30 languages, with human evaluations for a third of them, and enables participants without formal training to make a unique scientific contribution. Benchmarks, models, data, code, and evaluation results are released at \url{https://github.com/masakhane-io/masakhane-mt}.

\end{abstract}

\section{Introduction}\label{sec:intro}
Language prevalence in societies is directly bound to the people and places that speak this language. Consequently, resource-scarce languages in an NLP context reflect the resource scarcity in the society from which the speakers originate~\citep{digitaldivide}.
Through the lens of a machine learning researcher,  ``low-resourced'' identifies languages for which few digital or computational data resources exist, often classified in comparison to another language~\citep{gu-etal-2018-universal, zoph-etal-2016-transfer}. However, to the sociolinguist,  ``low-resourced'' can be broken down into many categories: low density, less commonly taught, or endangered, each carrying slightly different meanings~\citep{cieri-etal-2016-selection}. In this complex definition, the ``low-resourced''-ness of a language is a symptom of a range of societal problems, e.g. authors oppressed by colonial governments have been imprisoned for writing novels in their languages impacting the publications in those languages~\citep{wa1992decolonising}, or that fewer PhD candidates come from oppressed societies due to low access to tertiary education~\citep{britishcouncil}. This results in fewer linguistic resources and researchers from those regions to work on NLP for their language. Therefore, the problem of ``low-resourced''-ness relates not only to the available resources for a language, but also to the \emph{lack of geographic and language diversity of NLP researchers} themselves.

The NLP community has awakened to the fact that it has a diversity crisis in terms of limited geographies and languages~\citep{caines_2019,Joshi2020TheSA}: Research groups are extending NLP research to low-resourced languages~\cite{guzman2019flores,hu2020xtreme,wu2020all}, and workshops have been established~\citep{ws2018deep, winlp2019, emnlp-2019-deep}.

We scope the rest of this study to machine translation (MT) using parallel corpora only, and refer the reader to~\citet{joshi2019unsung} for an assessment of low-resourced NLP in general. 

\paragraph{Contributions.} We diagnose the problems of MT systems for low-resourced languages by reflecting on what agents and interactions are necessary for a sustainable MT research process. We identify which agents and interactions are commonly omitted from existing low-resourced MT research, and assess the impact that their exclusion has on the research. To involve the necessary agents and facilitate required interactions, we propose \emph{participatory research to build sustainable MT research communities for low-resourced languages}. The feasibility and scalability of this method is demonstrated with a case study on MT for African languages, where we present its implementation and outcomes, including novel translation datasets, benchmarks for over 30 target languages contributed and evaluated by language speakers, and publications authored by participants without formal training as scientists.

\begin{table}[t]
\resizebox{\columnwidth}{!}{%
    \centering
    \begin{tabular}{lrrl}
    \toprule
     \textbf{Language} & \textbf{Articles} & \textbf{Speakers} & \textbf{Category}\\
 \midrule
English & 6,087,118	& 1,268,100,000 & Winner \\ 
\midrule
Egyptian Arabic & 573,355	& 64,600,000 & Hopeful	\\
Afrikaans & 91,002	& 17,500,000 & Rising Star	\\
Kiswahili & 59,038	& 98,300,000 & Rising Star	\\
Yoruba & 32,572	& 39,800,000 & Rising Star	\\
Shona & 5,505	& 9,000,000	& Scraping by \\
Zulu & 2,219 & 27,800,000 & Hopeful \\
Igbo & 1,487	& 27,000,000 & Scraping by \\
Luo & 0 & 4,200,000 & Left-behind \\
Fon & 0 & 2,200,000 & Left-behind \\
Dendi & 0 & 257,000 & Left-behind\\
Damara & 0 & 200,000 & Left-behind \\
    \bottomrule
    \end{tabular}
    }
    \caption{Sizes of a subset of African language Wikipedias\footnote{\url{https://meta.wikimedia.org/wiki/List_of_Wikipedias}}, speaker populations\footnote{\url{https://www.ethnologue.com}}, and categories according to~ \citet{Joshi2020TheSA} (28 May 2020).}
    \label{tab:wikipedia_stats}
\end{table}

\section{Background} \label{sec:background}

\paragraph{Cross-lingual Transfer.} 
With the success of deep learning in NLP, language-specific feature design has become rare, and cross-lingual transfer methods have come into bloom~\citep{upadhyay-etal-2016-cross,ruder2019survey} to transfer progress from high-resourced to low-resourced languages~\citep{adams-etal-2017-cross,wang-etal-2019-cross,kim-etal-2019-effective}. 
The most diverse benchmark for multilingual transfer by~\citet{hu2020xtreme} allows measurement of the success of such transfer approaches across 40 languages from 12 language families. However, the inclusion of languages in the set of benchmarks is dependent on the availability of monolingual data for representation learning with previously annotated resources. The content of the benchmark tasks is English-sourced, and human performance estimates are taken from English. Most cross-lingual representation learning techniques are Anglo-centric in their design~\citep{anastasopoulos2019crosslingual}.

\paragraph{Multilingual Approaches.}  
Multilingual MT~\citep{dong-etal-2015-multi,firat-etal-2016-multi,firat-etal-2016-zero,wang2020balancing} addresses the transfer of MT from high-resourced to low-resourced languages by training multilingual models for all languages at once. ~\citep{aharoni-etal-2019-massively, massiveMTWild} train models to translate between English and 102 languages, for the 10 most high-resourced African languages on private data, and otherwise on public TED talks~\citep{qi-etal-2018-pre}.
Multilingual training often outperforms bilingual training, especially for low-resourced languages. However, with multilingual parallel data being also Anglo-centric, the capabilities to translate from English versus into English vastly diverge~\citep{zhang-etal-2020-improving}.

Another recent approach, mBART \citep{liu2020multilingual}, leverages both monolingual and parallel data and also yields improvements in translation quality for lower-resource languages such as Nepali, Sinhala and Gujarati.\footnote{Note that these languages have more digital resources available and a longer history of written texts than the low-resourced languages we are addressing here.}
While this provides a solution for small quantities of training data or monolingual resources, the extent to which standard BLEU evaluations reflect translation quality is not clear yet, since human evaluation studies are missing.

\paragraph{Targeted Resource Creation.}
\citet{guzman2019flores} develop evaluation datasets for low-resourced MT between English and Nepali, Sinhala, Khmer and Pashtolow. They highlight many problems with low-resourced translation: tokenization, content selection, and translation verification, illustrating increased difficulty translating from English into low-resourced languages, and highlight the ineffectiveness of accepted state-of-the-art techniques on morphologically-rich languages. Despite involving all agents of the MT process (Section~\ref{sec:process}), the study does not involve data curators or evaluators that understood the languages involved, and resorts to standard MT evaluation metrics. Additionally, how this effort-intensive approach would scale to more than a handful of languages remains an open question.

\section{The Machine Translation Process}
\label{sec:process}

\begin{figure}
    \centering
    \includegraphics[width=\columnwidth]{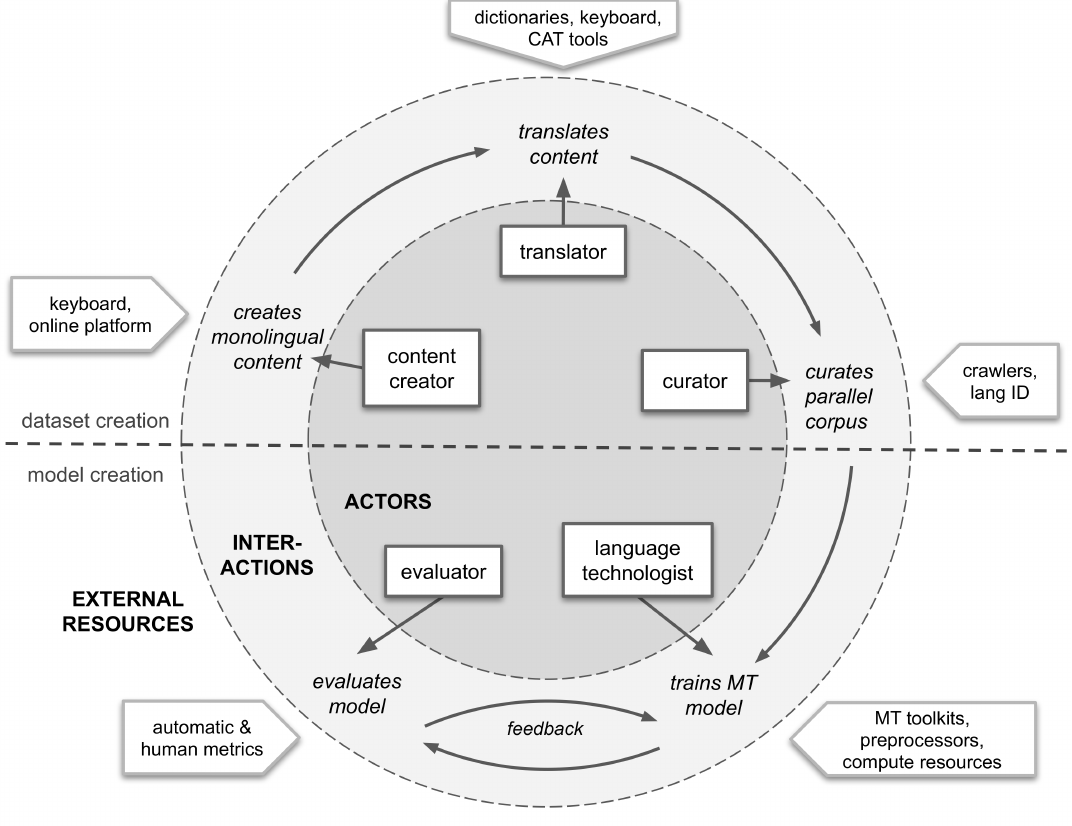}
    \caption{The MT Process, in terms of the necessary agents, interactions and external constraints and demand (excluding stakeholders).}
    \label{fig:diagram}
\end{figure}

We reflect on the process enabling a sustainable process for MT research on parallel corpora in terms of the required agents and interactions, visualized in Figure~\ref{fig:diagram}. 
Content creators, translators, and curators form the dataset creation process, while the language technologists and evaluators are part of the model creation process. Stakeholders (not displayed) create demand for both processes.

\textbf{Stakeholders} are people impacted by the artifacts generated by each agent in the MT process, and can typically speak and read the source or the target languages. 
To benefit from MT systems, the stakeholders need access to technology and electricity.

\textbf{Content Creators} produce content in a language, where content is any digital or non-digital representation of language. For digital content, content creators require keyboards, and access to technology.

\textbf{Translators} translate the original content, including crowd-workers, researchers, or translation professionals. They must understand the language of the content creator and the target language. A translator needs content to translate, provided by content creators. For digital content, the translator requires keyboards and technology access. 

\textbf{Curators} are defined  as individuals involved in the content selection for a dataset~\citep{bender2018data}, requiring access to content and translations. They should understand the languages in question for quality control and encoding information.

\textbf{Language Technologists} are defined as individuals using datasets and computational linguistic techniques to produce MT models between language pairs. Language technologists require language preprocessors, MT toolkits, and access to compute resources.

\textbf{Evaluators} are individuals who measure and analyse the performance of a MT model, and therefore need knowledge of both source and target languages. To report on the performance on models, evaluators require quality metrics, as well as evaluation datasets. Evaluators provide feedback to the Language Technologists for improvement. 

\subsection{Limitations of Existing Approaches}\label{sec:limitations}

If we place a high-resource MT pair such as English-to-French into the process defined above, we observe that each agent nowadays has the necessary resources and historical stakeholder demand to perform their role effectively. A ``virtuous cycle'' emerged where available content enabled the development of MT systems that in turn drove more translations, more tools, more evaluation and more content, which cycled back to improving MT systems.

By contrast, parts of the process for existing low-resourced MT are constrained.
Historically, many low-resourced languages had \emph{low demand from stakeholders} for content creation and translation~\cite{wa1992decolonising}. Due to \emph{missing keyboards or limited access to technology}, content creators were not empowered to write digital content~\cite{adam1997content,van2019writing}. This is a chicken-or-egg problem, where existing digital content in a language would attract more stakeholders, which would incentivize content creators~\cite{kaffee2018mind}. As a result, primary data sources for NLP research, such as Wikipedia, often have a few hundred articles only for low-resourced languages despite large speaker populations, see Table~\ref{tab:wikipedia_stats}. Due to limited demand, existing translations are often domain-specific and small in size, such as the JW300 corpus ~\citep{agic-vulic-2019-jw300} whose content was created for missionary purposes. 

When data curators are not part of the societies from where these languages originate, they are are often unable to identify data sources or translators for languages, prohibiting them from checking the \emph{validity of the created resource}. This creates problems in encoding, orthography or alignment, resulting in noisy or incorrect translation pairs~\cite{taylor2015data}. This is aggravated by the fact that many low-resourced languages do not have a long written history to draw from and therefore might be less standardized and using multiple scripts. In collaboration with content creators, data curators can contribute to \emph{standardization} or at least recognize potential issues for data processing further down the line.

As discussed in Section~\ref{sec:intro}, language technologists are fewer in low-resourced societies. Furthermore, the \emph{techniques developed in high-resourced societies might be inapplicable} due to compute, infrastructure or time constraints. Aside from the problem of education and complexity, existing techniques may not apply due to linguistic and morphological differences in the languages, or the scale, domain, or quality of the data~\citep{hu2020xtreme, pires-etal-2019-multilingual}.

Evaluators usually resort to potentially \emph{unsuitable automatic metrics} due to time constraints or missing connections to stakeholders~\cite{guzman2019flores}. The main evaluators of low-resourced NLP that is developed today typically cannot use human metrics due to the inability to speak the languages, or the lack of reliable crowdsourcing infrastructure, identified as one of the core weaknesses of previous approaches (in Section~\ref{sec:background}).

In summary, many agents in the MT process for low-resourced languages are either missing invaluable language and societal knowledge, or the necessary technical resources, knowledge, connections, and incentives to form interactions with other agents in the process.

\subsection{Participatory Research Approach}\label{sec:participatory}

We propose one way to overcome the limitations in Section~\ref{sec:limitations}: ensuring that \emph{the agents in the MT process originate from the countries where the low-resourced languages are spoken} or can speak the low-resourced languages. Where this condition cannot be satisfied, at least a \emph{knowledge transfer} between agents should be enabled. We hypothesize that using a participatory approach will allow researchers to improve the MT process by iterating faster and more effectively.

Participatory research, unlike conventional research, emphasizes the value of research partners in the knowledge-production process where the research process itself is defined collaboratively and iteratively. The ``participants'' are individuals involved in conducting research without formal training as researchers. Participatory research describes a broad set of methodologies, organised in terms of the level of participation. At the lowest level is crowd-sourcing, where participants are involved solely in data collection. The highest level---extreme citizen science--involves participation in the problem definition, data collection, analysis and interpretation~\citep{english2018crowdsourcing}.

Crowd-sourcing has been applied to low-resourced language data collection~\citep{ambati-etal-2010-active,guevara-rukoz-etal-2020-crowdsourcing,millour-fort-2018-toward}, but existing studies highlight how the disconnect between the data creation process and model creation process causes challenges. In seeking to create cross-disciplinary teams that emphasize the values in a societal context, a participatory approach which involves participants in every part of the scientific process appears pertinent to solving the problems for low-resourced languages highlighted in Section~\ref{sec:limitations}.

To show how more involved participatory research can benefit low-resource language translation, we present a case study in MT for African languages.

\section{Case Study: Masakhane}\label{sec:africanmt}

Africa currently has 2144 living languages~\citep{enthnologue}. Despite this, African languages account for a small fraction of available language resources, and NLP research rarely considers African languages.
In the taxonomy of \citet{Joshi2020TheSA}, African languages are assigned categories ranging from ``The Left Behinds'' to ``The Rising Stars'', with most languages not having any annotated data. Even monolingual resources are sparse, as shown in Table~\ref{tab:wikipedia_stats}.

In addition to a lack of NLP datasets, the African continent lacks NLP researchers. In 2018, only five out of the 2695 affiliations of the five major NLP conferences were from African institutions~\citep{caines_2019}. \citet{masakhaneIclr} attribute this to a culmination of circumstances, in particular their societal embedding~\citep{alexander2009evolving} and socio-economic factors, hindering participation in research activities and events, leaving researchers disconnected and distributed across the continent. Consequently, existing data resources are harder to discover, especially since these are often published in closed journals or are not digitized~\citep{mesthrie1995language}.

For African languages, the implementation of a standard crowd-sourcing pipeline as for example used for collecting task annotations for English, is at the current stage infeasible, due to the challenges outlined in Section~\ref{sec:process} and above. Additionally, no standard MT evaluation set for all of the languages in focus exists, nor are there prior published systems that we could compare all models against for a more insightful human evaluation. We therefore resort to intrinsic evaluation, and rely on this work becoming the \emph{first benchmark for future evaluations}.

We invite the reader to adopt a meta-perspective of this case study as an empirical experiment: Where the \emph{hypothesis} is that participatory research can facilitate low-resourced MT development; the \emph{experimental methodology} is the strategies and tools employed to bring together distributed participants, enabling each language speaker to train, contribute, and evaluate their models. The experiment is \emph{evaluated} in terms of the quantity and diversity of participants and languages, and the variety of research artifacts, in terms of benchmarks, human evaluations, publications, and the overall health of the community. While a set of novel human evaluation results are presented, they serve as demonstration of the value of a participatory approach, rather than the empirical focus of the paper. 

\subsection{Methodology}

To overcome the challenge of recruiting participants, a number of strategies were employed. Starting from local demand at a machine learning school (Deep Learning Indaba \citep{engelbrecht2018deep}), meetups and universities, distant connections were made through Twitter, conference workshops,\footnote{ICLR AfricaNLP 2020: \url{https://africanlp-workshop.github.io/}} and eventually press coverage\footnote{\url{https://venturebeat.com/2019/11/27/the-masakhane-project-wants-machine-translation-and-ai-to-transform-africa/}} and research publications.\footnote{\url{https://github.com/masakhane-io/masakhane-community/blob/master/publications.md}} To overcome the limited tertiary education enrollments in Sub-Saharan Africa~\citep{britishcouncil}, \emph{no prerequisites} were placed on researchers joining the project.
For the agents outlined in Section~\ref{sec:process}, no fixed roles are imposed onto participants. Instead, they join with a specific interest, background, or skill aligning them best to one or more of agents. To obtain cross-disciplinarity, we focus on the communication and interaction between participants to enable knowledge transfer between missing connections (identified in Section~\ref{sec:limitations}), allowing a fluidity of agent roles. For example, someone who initially joined with the interest of using machine translation for their local language (as a stakeholder) to translate education material, might turn into a junior language technologist when equipped with tools and introductory material and mentoring, and guide content creation more specifically for resources needed for MT.

To bridge large geographical divides, the community lives online. Communication occurs on GitHub and Slack with weekly video conference meetings and reading groups. Meeting notes are shared openly so that continuous participation is not required and time commitment can be organized individually. Sub-interest groups have emerged in Slack channels to allow focused discussions. Agendas for meetings and reading groups are public and democratically voted upon. In this way, the research questions evolve based on \emph{stakeholder demands}, rather than being imposed upon by external forces.

The lack of compute resources and prior exposure to NLP is overcome by providing tutorials for training a custom-size Transformer model with JoeyNMT~\citep{joey2019} on Google Colab\footnote{\url{https://colab.research.google.com}}. International researchers were not prohibited from joining. As a result, mutual mentorship relations emerged, whereby international researchers with more language technology experience guided research efforts and enabled data curators or translators to become language technologists. In return, African researchers introduced the international language technologists to African stakeholders, languages and context.

\subsection{Research Outcomes}

\paragraph{Participants.}
A growth to over 400 participants of diverse disciplines, from at least 20 countries, has been achieved within the past year, suggesting the participant recruitment process was effective. Appendix~\ref{app:demo} contains detailed demographics of a subset of participants from a voluntary survey in February 2020.
86.5\% of participants responded positively when asked if the community helped them find mentors or collaborators, indicating that the health of the community is positive. This is also reflected in joint research publications of new groups of collaborators.

\paragraph{Research Artifacts.} As a result of mentorship and knowledge exchange between agents of the translation process, our implementation of participatory research has produced artifacts for NLP research, namely datasets, benchmarks and models, which are publicly available online.\footnote{\url{ https://github.com/masakhane-io}}. 
Additionally, over 10 participants have gone on to publish works addressing language-specific challenges at conference workshops, such as \citep{dossou2020ffr,orife2020towards,orife2020improving,oktem2020tigrinya,van2020optimal,martinus2020neural,marivate2020investigating}.

\paragraph{Dataset Creation.}
The dataset creation process is ongoing, with new initiatives still emerging. We showcase a few initiatives below to demonstrate how bridging connections between agents facilitates the MT process.
\begin{enumerate}
    \item A team of Nigerian participants, driven by the internal demand to ensure that accessible and representative data of their culture is used to train models, are translating their own writings including personal religious stories and undergraduate theses into Yoruba and Igbo\footnote{\url{https://github.com/masakhane-io/masakhane-wazobia-dataset}}. 
    \item A Namibian participant, driven by a passion to preserve the culture of the Damara, is hosting collaborative sessions with Damara speakers, to collect and translate phrases that reflect Damara culture around traditional clothing, songs, and prayers.\footnote{ \url{https://github.com/masakhane-io/masakhane-khoekhoegowab}}
    \item Creating a connection between a translator in South-Africa's parliament and a language technologist has enabled the process of data curation, allowing access to data from the parliament in South-Africa's languages (which are public but obfuscated behind internal tools).\footnote{\url{http://bit.ly/raw-parliamentary-translations}}. 
\end{enumerate}
These stories demonstrate the value of including curators, content creators, and translators as participants. 

\paragraph{Benchmarks.} We publish 46 benchmarks for neural translation models from English into 39 distinct African languages, and from French into two additional languages, as well as from three different languages into English.\footnote{Benchmark scores can be found in Appendix~\ref{app:bleu}.} Most were trained on the JW300 corpus~\citep{agic-vulic-2019-jw300}. From this corpus, we select the English sentences most commonly found (and longer than 4 tokens) in all languages, as a global set of test sources. For individual languages, test splits are composed by selecting the translations that are available from this subset. While this biases the test set towards frequent segments, it prevents cross-lingual overlap between training and test data which has to be ensured for cross-lingual transfer learning. For training data, other sources like Autshumato~\citep{mckellar2014an}, TED~\citep{cettoloEtAl:EAMT2012}, SAWA~\citep{de-pauw-etal-2009-sawa}, Tatoeba\footnote{\url{https://tatoeba.org/}}, Opus~\citep{tiedemann2012parallel}, and data translated or curated by participants were added. Language pairs were selected based on the individual demands of each of the 33 participants, who voluntarily contributed the benchmarks they valued most. 16 of the selected target languages are categorized as ``Left-behind'' and 11 are categorized as ``Scraping by'' in the taxonomy of~\cite{Joshi2020TheSA}. The benchmarks are hosted publicly, including model weights, configurations and preprocessing pipelines for full reproducibility. The benchmarks are submitted by individual or groups of participants in form of a GitHub Pull Request. By this, we ensure that the contact to the benchmark contributors can be made, and ownership is experienced.

\subsection{Human MT Evaluation}
To our knowledge, there is no prior research on human evaluation specifically for machine translations of low-resourced languages. Until now, NLP practitioners were left with the hope that successful evaluation methodologies for high-resource languages would transfer well to low-resourced languages. This lack of study is due to the missing connections between the community of speakers (content creators and translators), and the language technologists. MT evaluations by humans are often done either within a group of researchers from the same lab or field (e.g. for WMT evaluations\footnote{\url{http://www.statmt.org/wmt19/}}), or via crowdsourcing platforms~\citep{ambati-vogel-2010-crowds,post-etal-2012-constructing}. Speakers of low-resource languages are traditionally underrepresented in these groups, which makes such studies even harder \citep{joshi2019unsung, guzman2019flores}.

One might argue that human evaluation should not be attempted before reaching a viable state of quality, but we found that early evaluation results in an improved  understanding of the individual challenges of the target languages, strengthens the network of the community, and most importantly, improves the connection and knowledge transfer between language technologists, content creators and curators. 

The ``low-resourced"-ness of the addressed languages pose challenges for evaluation beyond interface design or recruitment of evaluators proficient in the target language. For the example of Igbo, evaluators had to find solutions for typing diacritics without a suitable keyboard. In addition, Igbo has many dialects and variations which the MT model is uninformed of. Medical or technical terminology (e.g., ``data'') is difficult to translate and whether to use loan words required discussion. Target language news websites 
were found to be useful for resolving standardization or terminology questions.  Solutions for each language were shared and often also applicable for other languages.

\begin{table*}[htp!]
\resizebox{\textwidth}{!}{%
    \centering
    \begin{tabular}{lrccccccc}
    \toprule
        \textbf{Trg.} & \textbf{Train.} & \textbf{Autom.: JW300 }& \multicolumn{3}{c}{\textbf{Human: COVID}} & \multicolumn{3}{c}{\textbf{Human: TED}} \\
      \textbf{ lang.} &\textbf{ size }& \textbf{BLEU} $\uparrow$ & \textbf{HTER} $\downarrow$ &\textbf{ HBLEU} $\uparrow$& \textbf{HCHRF} $\uparrow$  & \textbf{HTER} $\downarrow$ & \textbf{HBLEU} $\uparrow$& \textbf{HCHRF} $\uparrow$\\ 
    \midrule
        \emph{ddn} & 6,937 & 22.30 & 1.11 & 0.27 & 0.08 & - &-  &- \\ 
        \emph{pcm} & 20,214 & 23.29 & 0.98 & 3.03 & 0.19 & 0.84 & 9.76 & 25.16\\ 
        \emph{fon} & 27,510 & 31.07 & 0.92 & 15.43 & 23.22 & - & - & - \\
        \emph{luo} & 136,459 & 34.33 & - & - & - & 1.26 & 7.90 & 20.88 \\ 
        \multirow{2}{*}{\emph{ha}} & \multirow{2}{*}{333,845} & \multirow{2}{*}{41.11} & 0.71 & 26.96 & 43.97 & 0.73 & 20.42 & 39.31 \\ 
         & &  & 0.64 & 26.56 & 46.71 & - & - & - \\ 
        \emph{ig} & 414,467 & 34.85 & 0.85 & 11.94 & 29.86 & 0.55 & 33.74 & 49.67 \\ 
        \emph{yo} & 415,100 & 38.62 & 0.09 & 85.92 & 89.90 & 0.51 & 49.22 & 58.41 \\ 
        \emph{sn} & 712,455 & 30.84 & 0.53 & 31.31 & 54.04 & - & - & -\\ 
        \emph{sw} & 875,558 & 48.94 & - & - & - & 0.32 & 60.47 & 78.67 \\ 
    \bottomrule
    \end{tabular} %
    }
    \caption{Evaluation results for translations from English. Metrics are computed based on Polyglot-tokenized translations.
    HTER are mean sentence-level TER scores computed with the Pyter Python package. BLEU and ChrF are computed with Sacrebleu and tokenize ``none''~\citep{post-2018-call}.}
    \label{tab:eval_results}    
\end{table*}

\paragraph{Data.} 
The models are trained on JW300 data.\footnote{Except for Hausa: multiple domains, see Table~\ref{tab:language_pairs}.} To gain real-world quality estimates beyond religious context, we assess the models' out-of-domain generalization by translating a English COVID-19 survey with 39 questions and statements regarding COVID-19,\footnote{\url{https://coronasurveys.org/}} where the human-corrected and approved translations can directly serve the purpose of gathering responses. The domain is challenging as it contains medical terms and new vocabulary. Furthermore, we evaluate a subset of the Multitarget TED test data~\citep{duh18multitarget}\footnote{\url{http://www.cs.jhu.edu/~kevinduh/a/multitarget-tedtalks/}}. The obtained translations enrich the TED datasets, adding new languages for which no prior translations exist. The size of the TED evaluations vary from 30 to 120 sentences. Details are given in Table~\ref{tab:eval_data}, Appendix~\ref{app:data}.

\paragraph{Evaluators.}
11 participants of the community volunteered to evaluate translations in their language(s), often involving family or friends to determine the most correct translations. The  evaluator role is therefore taken by both stakeholders and language technologists.
Within only 10 days, we gathered a total of 707 evaluated translations covering Igbo (\textit{ig}), Nigerian Pidgin (\textit{pcm}), Shona (\textit{sn}), Luo (\textit{luo}),  Hausa (\textit{ha}, twice by two different annotators), Kiswahili (\textit{sw}), Yoruba (\textit{yo}), Fon (\textit{fon}) and Dendi (\textit{ddn}). We did not impose prescriptions in terms of number of sentences to evaluate, or time to spend, since this was voluntary work, and guidelines or estimates for the evaluation of translations into these languages are non-existent.

\paragraph{Evaluation Technique.} Instead of a direct assessment~\citep{graham-etal-2013-continuous} often used in benchmark MT evaluations~\citep{barrault-EtAl:2019:WMT,guzman2019flores}, we opt for post-editing. Post-edits are grounded in actions that can be analyzed in terms of e.g. error types for further investigations, while direct assessments require expensive calibration \citep{bentivogli2018machine}.
Embedded in the community, these post-edit evaluations create an asset for the interaction of various agents: for the language technologists for domain adaptation, or for the content creators, curators, or translators for guidance in standardization or domain choice. 

\paragraph{Results.} Table~\ref{tab:eval_results} reports evaluation results in terms of BLEU evaluated on the benchmark test set from JW300, and human-targeted TER (HTER)~\citep{snover2006study}, BLEU \citep{papineni-etal-2002-bleu} and ChrF \citep{popovic-2015-chrf} against human corrected model translations. 
For \textit{ha} we find modest agreement between evaluators: Spearman's $\rho=0.56$ for sentence-BLEU measurements of the post-edits compared to the original hypotheses. 
Generally, we observe that the JW300 score is misleading, overestimating model quality (except \textit{yo}). Training data size appears to be a more reliable predictor of generalization abilities, illustrating the danger of chasing a single benchmark. However, \textit{ig} and \textit{yo} both have comparable amounts of training data, JW300 scores, and carry diacritics, but exhibit very different evaluation performances, in particular on COVID. This can be explained by the large variations of \textit{ig} as discussed above: Training data and model output are not consistent with respect to one dialect, while the \emph{evaluator} had to decide on one. We also find difference in performance across domains, with the TED domain appearing easier for \textit{pcm} and \textit{ig}, while the \textit{yo} model performs better on COVID.

\section{Conclusion}
We proposed a participatory approach as a solution to sustainably scaling NLP research to low-resourced languages. Having identified key agents and interactions in the MT development process, we implement a participatory approach to build a community for African MT. In the process, we discovered successful strategies for distributed growth and communication, knowledge sharing and model building. In addition to publishing benchmarks and datasets for previously understudied languages, we show how the participatory design of the community enables us to conduct a human evaluation study of model outputs, which has been one of the limitations of previous approaches to low-resourced NLP. The sheer volume and diversity of participants, languages and outcomes, and that for many for languages featured, this paper constitutes the first time that human evaluation of an MT system has been performed, is evidence of the value of participatory approaches for low-resourced MT. For future work, we will (1) continue to iterate, analyze and widen our benchmarks and evaluations, (2) build richer and more meaningful datasets that reflect priorities of the stakeholders, (3) expand the focus of the existing community for African languages to other NLP tasks, and (4) help implement similar communities for other geographic regions with low-resourced languages. 

\section*{Acknowledgements}

We would like to thank Benjamin Rosman and Richard Klein for their invaluable feedback, as well as the anonymous EMNLP reviewers, and George Foster and Daan van Esch. We would also like to thank Google Cloud for the grant that enabled us to build baselines for languages with larger datasets.

\bibliography{anthology,emnlp2020}
\bibliographystyle{acl_natbib}

\clearpage
\appendix

\section{Demographics}\label{app:demo}
\begin{figure}
    \centering
      \begin{subfigure}[b]{0.8\columnwidth}
    \centering
    \includegraphics[width=\columnwidth]{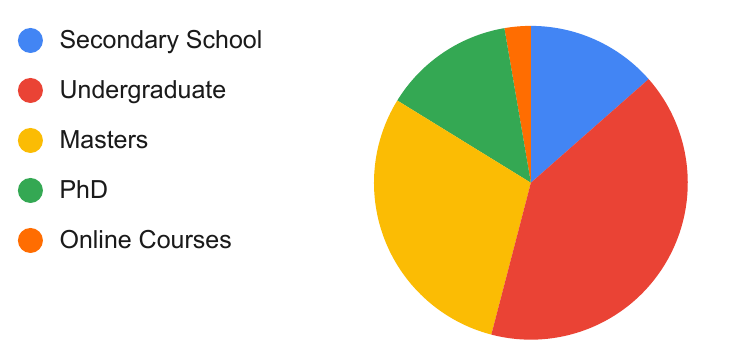}
    \caption{Highest Level of Education.}
    \label{fig:education}
    \includegraphics[width=\columnwidth]{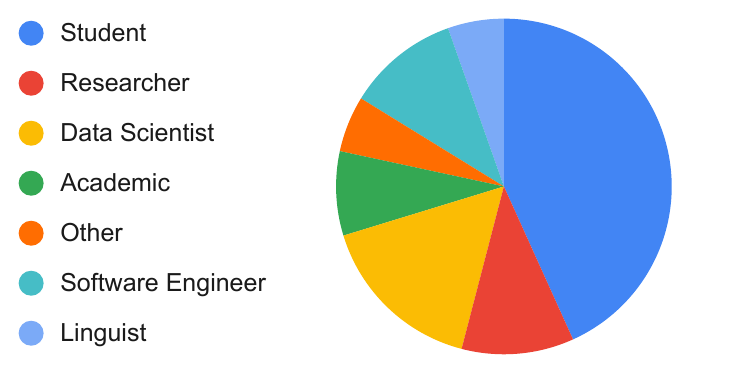}
    \caption{Occupation.}
    \label{fig:occupation}
    \end{subfigure}
    \caption{Education (a) and occupation (b) of a subset of 37 participants as indicated in a voluntary survey in February 2020.}
    \label{fig:demographics}
\end{figure}
Figure~\ref{fig:demographics} shows the demographics for a subset of participants from a voluntary survey conducted in February 2020. Between then and now (May 2020), the community has grown by 30\%, so these figures have to be seen as a snapshot. Nevertheless we can see that the educational background and the occupation is fairly diverse, with a majority of undergraduate students (not necessarily Computer Science).

\section{Evaluation Data}\label{app:data}

\begin{table}[t]
\centering
\begin{tabular}{lll}
    \toprule
     \textbf{Language} & \textbf{Domain} & \textbf{Size} \\
     \midrule
     Nigerian Pidgin & COVID &  39 \\
     & TED & 100 \\
     Luo & TED & 30 \\
     Yoruba & COVID & 39 \\
     & TED & 80 \\
     Hausa & COVID & 78 \\
     & TED & 120 \\
     Igbo & COVID & 39\\
     & TED & 50 \\
     Fon & COVID & 39 \\
     Swahili & TED & 55 \\
     Shona & COVID & 39 \\
     Dendi & COVID & 39 \\
     \bottomrule
\end{tabular}
\caption{Number of sentences for collected post-edits for TED talks and COVID surveys.}
\label{tab:eval_data}
\end{table}

Table~\ref{tab:eval_data} reports the number sentences that were post-edited in the human evaluation study reported in Section~\ref{sec:africanmt}.

\section{Benchmark Scores}\label{app:bleu}
Table~\ref{tab:language_pairs} contains BLEU scores on the JW300 test set for all benchmark models. BLEU scores are computed with Sacrebleu~\cite{post-2018-call} with tokenizer 'none' since the JW300 data comes tokenized with Polyglot.\footnote{\url{https://polyglot.readthedocs.io/en/latest/index.html}}. The table also features the target categories according to \cite{Joshi2020TheSA} as of 28 May 2020.

\begin{table*}[t]
    \centering
    \resizebox{0.7\textwidth}{!}{%
    \begin{tabular}{llll}
    \toprule

\textbf{Source} & \textbf{Target} & \textbf{Best Test BLEU} & \textbf{Category} \\
 \midrule
English & Afrikaans (Autshumato) & 19.56 &  Rising Star\\
English & Afrikaans (JW300) & 45.48 &  Rising Star\\
English & Amharic & 2.03 &  Rising Star\\
English & Arabic (TED, custom) & 9.28 & Underdog\\
English & Cinyanja & 30.00 & Scraping by\\
English & Dendi & 22.30 & Left Behind\\
English & Efik & 33.48 & Left Behind \\
English & \d{\`E}dó & 12.49 & Left Behind\\
English & \d{\`E}\d{\`s}án & 6.2 & Left Behind \\
English & Fon & 31.07 & Left Behind\\
English & Hausa (JW300+Tatoeba+more) & 41.11 & Hopeful\\
English & Igbo & 34.85 & Scraping by\\
English & Isoko & 38.91 & Left Behind\\
English & Kamba & 27.90 & Left Behind\\
English & Kimbundu & 32.76 & Left Behind\\
English & Kikuyu & 37.85 & Scraping by\\
English & Lingala & 48.64 & Scraping by\\
English & Luo & 34.33 & Left Behind\\
English & Nigerian Pidgin &  23.29 & Left Behind\\
English & Northern Sotho (Autshumato) & 19.56 & Scraping by\\
English & Northorn Sotho (JW300) & 15.40 & Scraping by\\
English & Sesotho  & 41.23 & Scraping by\\
English & Setswana &  19.66 & Hopeful\\
English & Shona & 30.84 & Scraping by\\
English & Southern Ndebele (I) &  4.01 & Left Behind\\
English & Southern Ndebele (II) & 26.61 & Left Behind \\
English & kiSwahili (JW300) & 48.94 &  Rising Star\\
English & kiSwahili (SAWA) & 3.60 &  Rising Star\\
English & Tigrigna (JW300) & 4.02 & Hopeful\\
English & Tigrigna (JW300+Tatoeba+more) & 14.88 & Hopeful\\
English & Tiv & 44.70 & Left Behind\\
English & Tshiluba & 42.52 & Left Behind\\
English & Tshivenda & 49.57 & Scraping by\\
English & Urhobo &  28.82 & Left Behind\\
English & isiXhosa (Autshumato) & 13.32 & Hopeful\\
English & isiXhosa (JW300) & 6.00 & Hopeful\\
English & Xitsonga (JW300) &  4.44 & Scraping by\\
English & Xitsonga (Autshumato) & 13.54 & Scraping by\\
English & Yoruba &  38.62 &  Rising Star\\
English & isiZulu (Autshumato) &  1.96 & Hopeful\\
English & isiZulu (JW300)&  4.87 & Hopeful\\
Efik & English & 33.68 & Winner\\
French & Lingala & 39.81 & Scraping by\\
French & Swahili Congo & 33.73 & Left Behind\\
Hausa & English & 25.27 & Winner\\
Yoruba & English &  39.44 & Winner\\
    \bottomrule 
    \end{tabular}
    }
    \caption{Benchmarks as of Nov 6, 2020. If not indicated, training domain is JW300. BLEU scores are computed with Sacrebleu (\textit{tokenize='none'}) on the JW300 test sets. Target languages are categorized according to \cite{Joshi2020TheSA} as of 28 May 2020.}
    \label{tab:language_pairs}
\end{table*}

\end{document}